%% file: main.tex
\documentclass[10pt,twocolumn,letterpaper]{article}

\usepackage{tabularx}
\usepackage{multirow} 
\usepackage{float} 
\usepackage{graphicx}
\usepackage{lipsum}
\usepackage{caption}

\usepackage[pagenumbers]{cvpr} 

\input{preamble}

\definecolor{cvprblue}{rgb}{0.21,0.49,0.74}
\usepackage[pagebackref,breaklinks,colorlinks,citecolor=cvprblue]{hyperref}

\def\paperID{26} 
\def\confName{CVPR}
\def\confYear{2024}

\title{KI-GAN: Knowledge-Informed Generative Adversarial Networks for Enhanced Multi-Vehicle Trajectory Forecasting at Signalized Intersections}

\author{Chuheng Wei$^1$ \qquad   Guoyuan Wu$^1$\qquad   Matthew J. Barth$^1$ \qquad Amr Abdelraouf$^2$ \\
\qquad Rohit Gupta$^2$ \qquad Kyungtae Han$^2$ \\
$^1$ University of California Riverside,  \qquad $^2$ InfoTech Labs, Toyota Motor North America\\
{\tt\small chuheng.wei@email.ucr.edu}
}

\begin{document}
\makeatletter
\g@addto@macro\@maketitle{
  \begin{figure}[H]
  \setlength{\linewidth}{\textwidth}
  \setlength{\hsize}{\textwidth}
  \centering
    \includegraphics[width=1\linewidth]{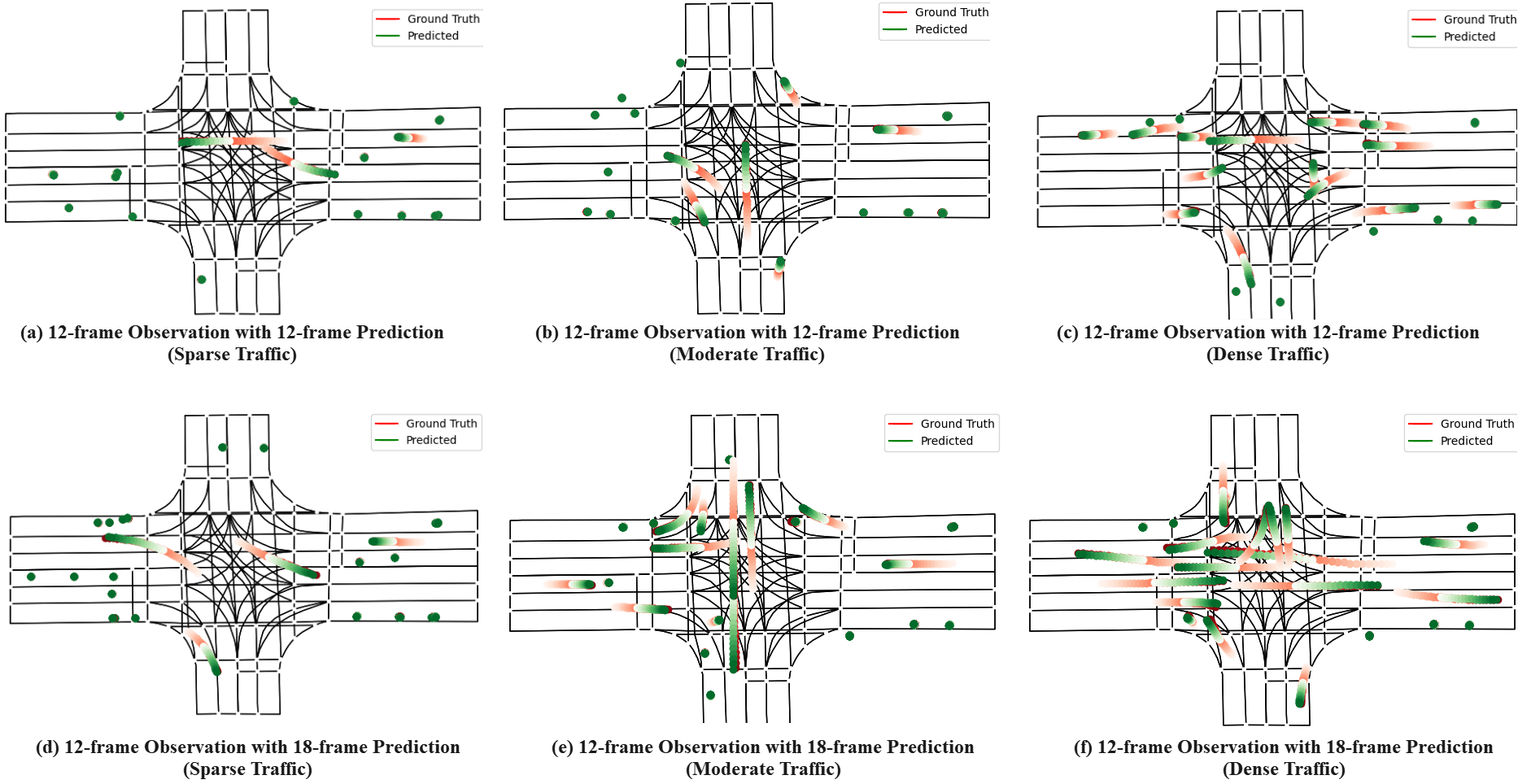}
    \caption{Trajectory Predictions by KI-GAN on SinD Dataset: Comparative Visualization of Ground Truth and Model Predictions. (Light red to dark red dots represent the trajectory's ground truth for both observed and predicted frames. Light green to dark green dots illustrate the KI-GAN model's predictions, highlighting accuracy and temporal progression.)}
    \label{fig:QualitativeResults}
  \end{figure}
}
\makeatother
\maketitle


\input{sec/0_abstract}

\input{sec/1_intro}
\input{sec/2_relatedwork}
\input{sec/3_methdology}
\input{sec/4_experiments}

\input{sec/5_conclusion}
{
    \small
    \bibliographystyle{ieeenat_fullname}
    \bibliography{main}
}


\end{document}

%% file: preamble.tex
\usepackage[dvipsnames]{xcolor}


%% file: sec/0_abstract.tex
\begin{abstract}

Reliable prediction of vehicle trajectories at signalized intersections is crucial to urban traffic management and autonomous driving systems. However, it presents unique challenges, due to the complex roadway layout at intersections, involvement of traffic signal controls, and interactions among different types of road users. To address these issues, we present in this paper a novel model called Knowledge-Informed Generative Adversarial Network (KI-GAN), which integrates both traffic signal information and multi-vehicle interactions to predict vehicle trajectories accurately. Additionally, we propose a specialized attention pooling method that accounts for vehicle orientation and proximity at intersections. Based on the SinD dataset, our KI-GAN model is able to achieve an Average Displacement Error (ADE) of 0.05 and a Final Displacement Error (FDE) of 0.12 for a 6-second observation and 6-second prediction cycle. When the prediction window is extended to 9 seconds, the ADE and FDE values are further reduced to 0.11 and 0.26, respectively. These results demonstrate the effectiveness of the proposed KI-GAN model in vehicle trajectory prediction under complex scenarios at signalized intersections, which represents a significant advancement in the target field.
Code is available at \href{https://github.com/ChuhengWei/KI_GAN}{$https://github.com/ChuhengWei/KI\_GAN$}.

\end{abstract}

%% file: sec/1_intro.tex
\section{Introduction}
\label{sec:intro}

In the era of urbanization, traffic management at signalized intersections has become increasingly crucial in order to maintain safety \cite{zhang2022d2}. These intersections serve as epicenters of vehicular movement, wherein the confluence of diverse road users and their interactions exacerbate traffic flow complexity \cite{Zyner2018Naturalistic}. Multiple vehicle trajectory forecasting in such environments is not only crucial for efficient traffic management systems, but also indispensable for reliable and effective autonomous driving \cite{Quintanar2021Predicting}. 
In this study, we introduce the \textit{Knowledge-Informed Generative Adversarial Network} (KI-GAN), a model specifically designed for predicting vehicle trajectories at signalized intersections. It is imperative to develop such a model to address the inherent challenges presented by urban traffic environments, such as intricate roadway layouts at intersections and traffic conditions and signal operations that are quite dynamic in nature.

\subsection{Background}
Trajectory prediction is a vital component in the development of efficient and safe transportation systems \cite{Quintanar2021Predicting}. In dynamic and interactive environments, forecasting the future path of vehicles based on their current and past states becomes increasingly challenging \cite{pellegrini2009you}. In the existing trajectory prediction research, linear scenarios, such as highways or non-interactive environments, have been primarily examined \cite{Dang2023Distributed,Zhan2018Towards,altche2017lstm}. However, with the growing complexity of urban traffic, particularly at intersections, the need for more sophisticated models has become evident \cite{Quintanar2021Predicting}. As opposed to highways and rural thoroughfares, intersections are characterized by their multidirectional flow patterns, divergent vehicular behavior, and traffic control signals, which distinguish them from other junctions \cite{eom2020traffic, zhao2022analysis}.

In multi-vehicle scenarios, the interaction between vehicles plays a crucial role \cite{Zhan2018Towards, abdelraouf2023interaction,li2023personalized}. Each vehicle's movement can influence the path of others, creating a dynamic system that is challenging to predict. In addition to physical interactions, drivers are also influenced by their perceptions, decisions, and reactions to surrounding environment, including roadway geometry\cite{mo2022multi}, other vehicles \cite{Roy2019Vehicle}, pedestrians \cite{Kawasaki2018Trajectory}, and traffic signs and signals \cite{Lee2023Deep}. These elements contribute to a certain level of complexity seldom encountered in simpler traffic scenarios. The status of traffic lights, for example, exerts a profound influence on driving behavior, prompting actions like acceleration, deceleration, or directional changes \cite{Fayazi2018Mixed-Integer, wei2024dilemma}. Consequently, anticipating vehicular paths in such environments becomes a notably intricate endeavor. Moreover, the heterogeneity in vehicle types and sizes introduces additional complexity layers \cite{Roy2019Vehicle}. For instance, larger vehicles like trucks exhibit distinct maneuvering and acceleration characteristics compared to smaller vehicles \cite{peng2023energy}, underscoring the need for a predictive model capable of assimilating and interpreting these diverse vehicular dynamics for precise trajectory forecasting.

In response to the complexities of intersection scenarios, our research integrates a wealth of intersection-specific information into our trajectory prediction model. We focus on the interactional layers by modeling essential vehicle characteristics such as speed, size, type, and their dynamics, including acceleration. Simultaneously, we incorporate a traffic light model, vital in influencing driver decisions and maneuvers. Additionally, recognizing the distinct nature of vehicle interactions at intersections, we introduce an innovative pooling method designed to capture the unique behavioral patterns of vehicles in these settings. This dual approach of detailed vehicle modeling and advanced interaction techniques marks a significant step towards more accurately predicting vehicle trajectories in the challenging environment of urban intersections.

\subsection{Contributions}
Addressing the challenges of the trajectory prediction at signalized intersection, our paper contributes significantly to the following fields:

\textbf{Introduction of KI-GAN: } Our primary contribution is the development of Knowledge-Informed Generative Adversarial Networks (KI-GAN), a model that stands out for its ability to integrate diverse and critical data sources. By assimilating diverse vehicle data, traffic signal information, and multi-vehicle interactions, KI-GAN offers a comprehensive understanding of the factors influencing vehicle trajectories at intersections. The core of its innovation lies in the multi-module encoder, a feature that allows the model to process and interpret various types of input data effectively. This encoder is essential for capturing the complex, multi-layered interactions and conditions typical of urban intersections, thereby enabling the model to predict vehicle trajectories with a higher degree of accuracy and reliability.

\textbf{Specialized Attention Pooling Method: } 
Complementing the KI-GAN, we introduce an advanced attention pooling method specifically designed for analyzing vehicle orientation and proximity in intersection scenarios. This method, known as Vehicle Attention Pooling Net (VAP-Net), marks a significant improvement over traditional interaction modules used in trajectory prediction models. It is adept at discerning the subtle yet critical aspects of vehicle behavior and interaction, such as speed and orientation towards other vehicles and response to nearby traffic elements. Our empirical results demonstrate the superiority of VAP-Net in accurately capturing and predicting the nuanced movements of vehicles in complex urban intersection environments.

\begin{figure*}[ht]
    \centering
    \includegraphics[width=1\linewidth]{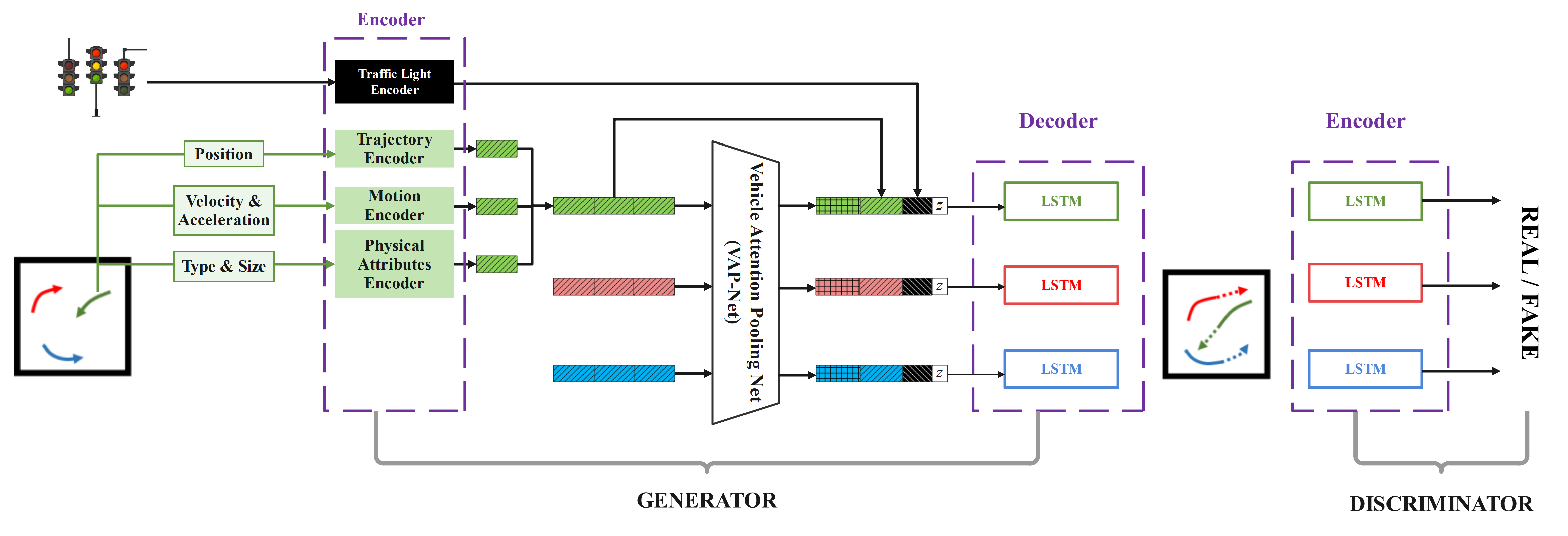}
    \caption{Architecture of Knowledge-Informed Generative Adversarial Networks}
    \label{fig:architecture}
\end{figure*}

%% file: sec/2_relatedwork.tex
\section{Related Work}

In this review, we focus on trajectory prediction methods tailored for the unique dynamics of signalized intersections and the integral multi-vehicle interactions. Signalized intersections require sophisticated prediction methods due to variables like traffic signals and vehicle maneuvers, while interaction-based multi-vehicle predictions are crucial for a realistic depiction of collective traffic behavior. 

\subsection{Trajectory Prediction Models for Intersection Scenarios}
Some researchers have attempted to tailor deep learning algorithms specifically for trajectory prediction at signalized intersections. Abdelraouf et al. \cite{abdelraouf2022trajectory} used sequence-to-sequence Long Short-Term Memory (LSTM) networks for intersection-wide vehicle trajectory prediction. Their proposed algorithm predicted future vehicle positions and orientations. 
Kawasaki and Tasaki \cite{Kawasaki2018Trajectory} designed a model specifically for predicting turning trajectories at intersections, considering the geometry of the crossing and the speeds of observed vehicles. Cao et al. \cite{Cao2021Real-Time} developed a vision-based intersection trajectory prediction model that predicts vehicle maneuvers such as left turns, right turns, or going straight, aiming to prevent traffic collisions. While these models mark progress in intersection-specific prediction, they often overlook the critical role of traffic signals.

As some trial efforts, Roy et al. \cite{Roy2019Vehicle} incorporated vehicle size in their (GAN)-based approach for both signalized and non-signalized intersections, but without directly considering traffic signal status. Oh and Peng \cite{Oh2019Impact} focused on the impact of traffic lights on vehicle speed rather than trajectory variation. Zhang et al.'s D2-TPred model \cite{zhang2022d2}, using a spatial dynamic interaction graph (SDG) and a behavior dependency graph (BDG), is a notable exception, as it considered traffic light-induced trajectory discontinuities. However, these models still tend to overlook a comprehensive view of vehicle characteristics in their interaction-based trajectory predictions.

In conclusion, while there has been significant progress in trajectory prediction at intersections, existing models often fall short in simultaneously addressing the complex vehicle interactions and the impact of traffic signals. This gap highlights the need for a comprehensive approach like the proposed KI-GAN model, which integrates these critical aspects for enhanced trajectory forecasting at signalized intersections.

\subsection{Multi-Vehicle Trajectory Prediction Considering Interactions}
Forecasting trajectories at intersections is complex due to significant vehicle interactions and dense traffic conditions. This complexity is mirrored in pedestrian trajectory prediction research \cite{social-gan,socialpooling,kosaraju2019social}, where the interplay between multiple agents has also been recognized as a pivotal element. In interactive scenarios, vehicle accidents are riskier than pedestrian incidents, requiring analysis of more complex data including vehicle size, type, and speed, along with historical trajectory 
Early studies, like those by Pecher et al. \cite{Pecher2016Data-Driven}, recognized the importance of considering neighboring vehicles' trajectories, laying the groundwork for more complex interaction models. Building on this, Lin et al. \cite{Lin2022Vehicle} introduced the concept of Spatio-Temporal Attention in their Long Short-Term Memory (STA-LSTM) model to better understand the influence of historical and nearby vehicle trajectories. Similarly, Mo et al. \cite{Mo2021Graph} explored the social interaction behaviors among vehicles, especially on highways, highlighting how the number of vehicles can influence behaviors and trajectories.

In the realm of multi-vehicle interactions, models have become increasingly sophisticated. 
Salzmann et al. \cite{salzmann2020trajectron++} introduced the modular, graph-structured Trajectron++ model, which predicts the trajectories of multiple agents by integrating agent dynamics and semantic maps.
Subsequently, other scholars \cite{Mersch2021Maneuver-based,navarro2022social, liu2022multi, liu2021multimodal,rowe2023fjmp} advanced this by predicting multiple potential behaviors and future positions of surrounding vehicles, using convolutional neural networks and grid map predictions, respectively.

However, most of these studies did not fully account for the unique complexities at intersections, such as the intricacies of roadway layout, vehicle orientation, and diverse vehicle types. These factors necessitate a multi-parameter model to capture the intricate dynamics at intersections, where the influence of nearby vehicles varies based on proximity and speed.

\subsection{Research Gaps}
Based on a comprehensive review of existing research, we have found the following key research gaps in trajectory prediction for intersections:

\textbf{Traffic Light Influence:} Most current models do not adequately factor in the influence of traffic light status on vehicle behavior at intersections. Traffic light status has been shown to have significant effect over driving behaviour at intersections \cite{islam2023effect}. This oversight can lead to significant inaccuracies in trajectory prediction.

\textbf{Multidimensional Data Integration:} Traditional models often rely on simplistic representations of vehicle interactions, primarily based on positional data. However, at intersections, the integration of additional data types such as vehicle speed \cite{messaoud2020attention}, size \cite{dey2017impact}, type \cite{Roy2019Vehicle}, and traffic signal status \cite{zhang2022d2} is crucial for a realistic and accurate prediction model.

\textbf{Advanced Interaction Pooling:} Existing models lack a specialized mechanism to handle the complex interactions at intersections. A new approach that can effectively pool interaction data, taking into account the diverse factors influencing vehicle behavior at intersections, is needed.

%% file: sec/3_methdology.tex
\section{Methodology}

\subsection{Framework of KI-GAN}
The Knowledge-Informed Generative Adversarial Network (KI-GAN) framework is presented in this section, a sophisticated model designed to accurately predict vehicular traffic patterns. Figure \ref{fig:architecture} illustrates our model's architecture, where the trajectory of a single agent (highlighted in green) is fully processed through multiple encoders to form a combined feature vector. It should be noted that the construction of the combined feature vector for other agents in the scene adheres to the same methodology, ensuring a unified approach to trajectory prediction across all agents.
As illustrated in the figure, our framework employs a multi-encoder approach, integrating diverse data streams to capture the complex dynamics of vehicle movement and interaction. This framework allows the generation of realistic and socially-aware trajectory predictions through the structured synthesis of spatial, vehicular, and traffic information.
\subsubsection{Encoders} The generator consists of four specialized encoders: Trajectory Encoder, Physical Attributes Encoder, Motion Encoder,and Traffic Encoder, each designed to process distinct types of input data.
\begin{itemize}
    \item \textbf{Trajectory Encoder:} This module focuses on spatial coordinates (x, y) of vehicles. Using Recurrent Neural Network (RNN) architectures, it captures temporal dependencies and movement patterns in trajectory data, offering insight into the vehicles' historical paths.
    \begin{equation}
        E_{\text{traj}} = \text{RNN}_{\text{traj}}(X, Y),
    \end{equation}
    where $X$ and $Y$ refer to  vectors composed of coordinates of trajectory points.
    \item\textbf{Motion Encoder:} This encoder deals with dynamic vehicle states, including velocity and acceleration. Similar to the Trajectory Encoder, it also utilizes an RNN to understand the temporal evolution of these state variables, providing a real-time context to the vehicle's movement.
    \begin{equation}
        E_{\text{motion}} = \text{RNN}_{\text{state}}(V, A),
    \end{equation}
    where $V$  stands for a vector composed of vehicle speeds, and $A$  represents a vector composed of vehicle accelerations.
    \item\textbf{Physical Attributes Encoder:} This part of the Generator encodes vehicle-specific information such as type and dimensions. By employing separate embedding layers for categorical data (like vehicle type) and linear layers for continuous data (dimensions), it creates a comprehensive feature vector for each vehicle.
    
    \begin{equation}
        E_{\text{phy}} = \text{Concat}(\text{Embed}_{\text{type}}(T), \text{Linear}_{\text{size}}(L, W)),
    \end{equation}
    where $T$ denotes the type of the vehicle, including cars, trucks, buses, etc., and $L$ and $W$ indicate the length and width of the vehicle from a top-down perspective. 
    \item\textbf{Traffic Encoder:} Dedicated to processing the traffic state information, this encoder uses an embedding layer to transform discrete traffic states into a higher-dimensional space. For example, we encode the status of traffic lights at intersections in the SinD dataset \cite{sind} using numbers 1 to 5 as shown in Table \ref{tab:trafficEncoder}. This transformation allows the model to encode and interpret various traffic conditions effectively.
    \begin{equation}
        E_{\text{traffic}} = \text{Embed}_{\text{traffic}}(T_{\text{state}}),
    \end{equation}
    where $T_{\text{state}}$ signifies the traffic light status at the intersection.
\end{itemize}
\begin{table}[htbp]
\caption{Traffic light status in each direction and the corresponding digital codes (The SinD dataset features an intersection with a single traffic light per direction, lacking dedicated left-turn signals, resulting in a unified signal state per direction. Moreover, the east-west and north-south traffic light states are synchronized respectively.)}
\begin{center}
\begin{tabular}{cccc|c}
\hline
\multicolumn{4}{c}{\textbf{Direction}} &\\
North & South & East & West &  Code\\
\midrule
Green & Green & Red & Red & 1\\
Yellow & Yellow & Red & Red & 2\\
Red & Red & Red & Red & 3\\
Red & Red & Green & Green & 4\\
Red & Red & Yellow & Yellow & 5\\
\bottomrule
\end{tabular}
\label{tab:trafficEncoder}
\end{center}
\end{table}
\subsubsection{Social Interaction} After encoding the different data streams, the features from the Trajectory, Motion, and Physical Attributes Encoders are concatenated to form a combined feature vector. 
\begin{equation}
    F_{\text{combined}} = \text{Concat}(E_{\text{traj}}, E_{\text{motion}}, E_{\text{phy}}).
\end{equation}
This vector, representing each trajectory, is then fed into a pooling layer along with features from other trajectories. The pooling layer aggregates contextual information from neighboring trajectories, enabling the model to understand the social and spatial interactions among vehicles. For the pooling layer, we propose a \textbf{Vehicle Attention Pooling Net (VAP-Net)}, which is designed to effectively calculate and interpret potential interactions and interrelations between vehicle trajectories.
\begin{equation}
    P = \text{VAPNet}(F_{\text{combined}}, (X,Y), V),
\end{equation}
where $(X, Y)$ denote vectors composed of trajectory coordinates, and $V$ indicates a vector composed of speeds.
\subsubsection{Decoder}
The output of the pooling layer, when amalgamated with the encoded traffic state and the combined feature vector, constitutes the comprehensive input to the decoder. This integration is mathematically represented as:
\begin{equation}
    F_{\text{recombined}} = \text{Concat}(P, F_{\text{combined}}, E_{\text{traffic}}).
\end{equation}
The decoder is realized through an LSTM \cite{lstm} framework, which utilizes the enriched input \( F_{\text{recombined}} \) to forecast future vehicle trajectories. The LSTM, with its recurrent architecture, is adept at generating sequences that are temporally consistent and socially acceptable by reflecting on historical data, present ego-vehicle states, and the surrounding traffic conditions\cite{lstm}. This dynamic is encapsulated in the equation:
\begin{equation}
    \text{Trajectory}_{\text{future}} = \text{LSTM}([F_{\text{recombined}}; z]).
\end{equation}
Here, \( z \) introduces a degree of randomness, enabling the model to generate a spectrum of possible future trajectories, contributing to its robustness. As a result of the blending of these multifaceted data streams, the model offers a comprehensive insight into vehicular dynamics, paving the way for reliable and precise predictions. This model's versatility is attributed to its ability to synthesize such a wide range of predictive capabilities across a variety of traffic conditions.

\subsubsection{Discriminator} The Discriminator assesses the generated paths by comparing them against actual vehicle movements, assigning a score that reflects their authenticity \cite{Zhao2021A}. This evaluation not only categorizes the paths as genuine or fabricated but also enhances the fidelity of the trajectory generation by providing feedback to the Generator.

The Discriminator's objective is formalized within the GAN's loss function, where it seeks to maximize its classification accuracy, thus enabling a refined generation of trajectories. Simultaneously, the model employs an L2 loss to quantify the deviation of the generated paths from the true trajectories, thereby selecting the most plausible path among multiple predictions.

Mathematically, the Discriminator's process is represented as follows:

Objective Function of GAN:
\begin{align}
\min_{G} \max_{D} V(G, D) &= \mathbb{E}_{x \sim p_{data}(x)}[\log D(x)] \nonumber \\
&+ \mathbb{E}_{z \sim p_{z}(z)}[\log (1 - D(G(z)))].
\end{align}

L2 Loss Metric:
\begin{equation}
L = \min_{k} \| Y_{i} - \hat{Y}_{i}^{(k)} \|_{2},
\end{equation}

where, \(G\) denotes the Generator, \(D\) the Discriminator, \(x\) the real trajectory data, \(z\) the input noise vector, \(Y_{i}\) the actual trajectory, and \(\hat{Y}_{i}^{(k)}\) the k-th predicted trajectory by the Generator.

\subsection{Vehicular Attention Pooling Net (VAP-Net)}

Generative Adversarial Networks (GANs) \cite{goodfellow2014generative} employ a pooling layer that is vital for managing information sharing among objects in a scene. This layer calculates agents' relative positions and merges their hidden states to represent their interactions. For vehicle trajectory prediction at intersections, traffic light status is a key factor influencing driver behavior \cite{Lee2023Deep}, and the proximity between vehicles and to the stop-bar largely dictates driving decisions \cite{dey2017impact}. Additionally, at intersections, drivers must consider the speed \cite{Kawasaki2018Trajectory} and direction \cite{atten-GAN} of nearby vehicles. Faster vehicles are monitored more closely due to their higher potential for danger, and vehicles from opposing directions are also critical to note from the collision risk perspective.




Research has shown that vehicle interactions at intersections involve complexities beyond mere positional relationships, as the dynamics of surrounding vehicles, including their speed and direction, significantly influence decision-making during interactions \cite{Kim2017Probabilistic, Dang2023Distributed,ye2021tpcn}.
Considering the particular nature of vehicle trajectory prediction at intersections, we present a novel pooling network architecture known as \textit{Vehicular Attention Pooling Net (VAP-Net)}. By incorporating a mechanism of attention in addition to positional information into VAP-Net, the model enhances its ability to capture dynamic interactions between complex agents. By integrating various streams of information, including positional information, velocity vectors, and historical information, it is designed to intricately process and integrate information. The architecture effectively combines spatial and velocity embeddings with an attention-driven pooling mechanism to create nuanced representations of agent interactions.

\subsubsection{Embedding and Interaction Encoding}

VAP-Net begins its operation by embedding spatial and velocity information for each agent in the scene:

\paragraph{Relative Position Embedding}
For each pair of agents \( i, j \), the relative position vector \( \mathbf{RP}_{ij} = \mathbf{p}_i - \mathbf{p}_j \) is computed, where \( \mathbf{p}_i \) and \( \mathbf{p}_j \) denote their respective positional coordinates. This vector is then transformed through a spatial embedding layer:
\begin{equation}
\mathbf{E}_{ij} = f_{\text{spatial}}(\mathbf{RP}_{ij}),
\end{equation}
where \( f_{\text{spatial}} \) represents the spatial embedding function.

\paragraph{Velocity Vector Embedding}
Simultaneously, the velocity vector \( \mathbf{v}_i \) of each agent is embedded:
\begin{equation}
\mathbf{V}_i = f_{\text{velocity}}(\mathbf{v}_i),
\end{equation}
with \( f_{\text{velocity}} \) being the velocity embedding function.

\subsubsection{Attention Mechanism}

The attention mechanism dynamically assigns weights to features based on their relevance:

\paragraph{Attention Weight Calculation}
For each agent pair, we concatenate their relative position and velocity embeddings and input this to an attention MLP. The output attention scores are normalized to form weights:
\begin{equation}
\mathbf{W}_{ij} = \text{Softmax} \left( f_{\text{attention}}([\mathbf{E}_{ij}; \mathbf{V}_i]) \right),
\end{equation}
where \( f_{\text{attention}} \) denotes the attention MLP.

\subsubsection{Feature Integration and Contextual Pooling}

\paragraph{Feature Aggregation}
The agent's hidden states \( \mathbf{H}_i \) are modulated by the attention weights to emphasize significant interactions:
\begin{equation}
\mathbf{H}'_{ij} = \mathbf{H}_i \odot \mathbf{W}_{ij},
\end{equation}
where \( \odot \) signifies element-wise multiplication.

\paragraph{Contextual Pooling}
The weighted hidden states are then concatenated with the relative position embeddings. This concatenated vector is processed through a series of MLPs, and a max-pooling operation is applied to extract a refined representation. We define this output as the Attention Pooling Feature (\( \mathbf{APF}_{i}\)), which encapsulates the fused characteristics of individual movement and interactive dynamics:
\begin{equation}
\mathbf{APF}_i = \max \left( \mathbf{MLP} \left( [\mathbf{E}_{ij}; \mathbf{H}'_{ij}] \right) \right),
\end{equation}
capturing both individual motion characteristics and interaction dynamics.

Through the integration of Pool Hidden Net's feature processing with an advanced attention mechanism, VAP-Net provides a context-aware synthesis of multi-agent interactions. This representation is pivotal for accurate prediction of future trajectories in complex and dynamic environments.

%% file: sec/4_experiments.tex
\section{Experiments}

\subsection{Dataset}


In the development of our proposed KI-GAN algorithm, a substantial set of vehicular movement parameters is required to accurately simulate and predict the interactions at intersections. These parameters include not only the basic metrics such as the positions, speeds, and accelerations of the vehicles but also extend to more detailed aspects like the vehicle type and size, as well as the status of traffic lights in each direction within the intersections. After a thorough comparison and analysis of a range of available datasets\cite{zhan2019interaction,bock2020ind,chang2019argoverse, wang2020ethical,krajewski2020round, breuer2020opendd}, we found that the SinD dataset \cite{sind} stands out as the only dataset that comprehensively meets all our requirements for input data. Its extensive coverage of the aforementioned parameters makes it exceptionally suitable for the verification algorithm in this article, offering the precise blend of data necessary to support the intricate computations and simulations at the heart of our research.



The SinD dataset, originating from a signalized intersection in Tianjin, China, stands as a rich and detailed resource for trajectory analysis, encompassing over 13,000 traffic participants across seven categories, including cars, buses, trucks, motorcycles, bicycles, tricycles, and pedestrians, over a span of seven hours. Besides providing comprehensive information regarding the type, dimensions, and trajectory of each vehicle, this dataset also provides comprehensive data regarding traffic light states and timings, which are crucial for studies related to vehicle dynamics and compliance with traffic signals. Its emphasis on signalized intersections and diverse traffic participant categories, offering a wealth of detailed information, meets the specific needs of our model for comprehensive urban traffic analysis.

\subsection{Experimental Setup and Evaluation Metrics}

The original SinD dataset captures data at 30 frames per second. For our study, we sampled the data at a step size of 15, resulting in a frequency of 2 frames per second. In situations of minimal vehicle movement or complete stops at intersections influenced by traffic signals, brief observation intervals can result in skewed data, which can negatively affect the model's ability to learn dynamic movements. Thus, we observed data for 6 seconds (12 frames) and predicted trajectories for both 6 seconds (12 frames) and 9 seconds (18 frames).

Regarding hyperparameters, we selected a batch size of 64 and trained the model for 50 epochs. The learning rate for the Generator was set to 0.001, and for the Discriminator, it was set to be 0.0005.

We used Average Displacement Error (ADE) and Final Displacement Error (FDE) as metrics for trajectory accuracy evaluation. These are defined as:

\begin{equation}
    \text{ADE} = \frac{1}{N}\sum_{i=1}^{N} \frac{1}{T} \sum_{t=1}^{T} \sqrt{(x_{t}^i - \hat{x}_{t}^i)^2 + (y_{t}^i - \hat{y}_{t}^i)^2},
\end{equation}

\begin{equation}
    \text{FDE} = \frac{1}{N}\sum_{i=1}^{N} \sqrt{(x_{T}^i - \hat{x}_{T}^i)^2 + (y_{T}^i - \hat{y}_{T}^i)^2},
\end{equation}

where $N$ is the total number of trajectories, $T$ the prediction horizon, $(x_t^i, y_t^i)$ the ground truth coordinates, and $(\hat{x}_t^i, \hat{y}_t^i)$ the predicted coordinates at time $t$ for the $i$-th trajectory.

Moreover, experiments for this study were performed using an Nvidia RTX 3090 graphics card, with the code developed in Pytorch.

\subsection{Results}

\textbf{Quantitative Results}

The SinD dataset encompasses 23 segments, out of which 20 were allocated for training and 3 for validation. Table \ref{tab:results} presents the evaluation metrics on the validation set, showcasing the results for both 12-frame observations with 12-frame predictions and 12-frame observations with 18-frame predictions. 
\begin{table}[ht]
\caption{Evaluation Metrics for Trajectory Prediction at Signalized Intersection \begin{small}
(*In adapting the algorithm from \cite{Lee2023Deep} for SinD dataset, which lacks public video information, we have omitted the vision encoding module. Instead, we focus on employing trajectory data, vehicle states, and traffic light status for our comparison models, aligning with the available dataset features. )
\end{small}
}
\begin{center}
\begin{tabular}{ccccc}
\toprule
\multirow{2}{*}{\textbf{Method}} & \multicolumn{2}{c}{\textbf{12-12}} & \multicolumn{2}{c}{\textbf{12-18}} \\
\cmidrule(lr){2-3} \cmidrule(lr){4-5} 
& \textbf{ADE} & \textbf{FDE} & \textbf{ADE} & \textbf{FDE} \\
\midrule
S-GAN\cite{social-gan} & 1.32  & 2.46  & 1.53  & 2.95  \\
S-LSTM\cite{socialpooling} & 0.87  & 1.60  & 0.96  & 1.78  \\

Trajetron++\cite{salzmann2020trajectron++} & 0.37  & 0.93  & 0.70  & 1.91  \\

FJMP \cite{rowe2023fjmp} & 0.27  & 0.68  & 0.41  & 1.13  \\

\cite{Lee2023Deep}* & 0.10  & 0.21  & 0.16  & 0.38  \\
\textbf{KI-GAN} & \textit{0.05} & \textit{0.12} & \textit{0.11} & \textit{0.26}\\
\hline
\end{tabular}
\label{tab:results}
\end{center}
\end{table}

\begin{table*}[h]
\caption{Impact of Encoder Components on Trajectory Prediction Accuracy}
\centering
\small
\begin{tabularx}{1\linewidth}{cccccccc}
\toprule
\multicolumn{3}{c}{Interaction Encoder}& \multirow{3}{*}{\parbox{1cm}{Traffic\\ Encoder}} & \multicolumn{2}{c}{12-12} & \multicolumn{2}{c}{12-18} \\ 
\cmidrule(lr){1-3} \cmidrule(lr){5-6} \cmidrule(lr){7-8} 
Trajectory & Motion & Physical Attri- &  & \multirow{2}{*}{ADE} & \multirow{2}{*}{FDE} & \multirow{2}{*}{ADE} & \multirow{2}{*}{FDE} \\ 
Encoder & Encoder & bute Encoder &  &  &  &  &  \\ 
\midrule
$\checkmark$ & $\checkmark$  & $\checkmark$ & $\times$  & 0.52 ($\uparrow$ 940.0 \%) & 1.13 ($\uparrow$ 841.7 \%) & 0.79 ($\uparrow$ 618.2 \%) & 1.54 ($\uparrow$ 492.3 \%) \\ 
$\checkmark$  & $\checkmark$ & $\times$   & $\checkmark$   & 0.14 ($\uparrow$ 180.0 \%) & 0.32 ($\uparrow$ 166.7 \%) & 0.22 ($\uparrow$ 100.0 \%) & 0.48 ($\uparrow$ 84.6 \%) \\   
$\checkmark$    & $\times$  & $\checkmark$  & $\checkmark$   & 0.12 ($\uparrow$ 140.0 \%) & 0.27 ($\uparrow$ 125.0 \%) & 0.17 ($\uparrow$ 54.5\%) & 0.37 ($\uparrow$ 42.3 \%)\\
$\checkmark$ & $\checkmark$ & $\checkmark$  & $\checkmark$   & 0.05 & 0.12 & 0.11 & 0.26 \\ 
\bottomrule
\end{tabularx}
\label{abla1}
\end{table*}

\begin{table*}[h]
\caption{Comparison of Pooling Methodologies on Trajectory Prediction Performance}
\centering
\begin{tabular}{lcccc}
\toprule
Pooling Method & \multicolumn{2}{c}{12-12} & \multicolumn{2}{c}{12-18} \\ 
\cmidrule(lr){2-3} \cmidrule(lr){4-5} 
               & ADE & FDE & ADE & FDE \\ 
\midrule
Social Pool Net\cite{socialpooling} & 0.07 ($\uparrow$ 40.0 \%) & 0.15 ($\uparrow$ 25.0 \%) & 0.14 ($\uparrow$ 27.3 \%)  & 0.33 ($\uparrow$ 26.9 \%)\\ 
Hidden Pool Net\cite{social-gan} & 0.08 ($\uparrow$ 60.0 \%) & 0.16 ($\uparrow$ 33.3 \%) & 0.19 ($\uparrow$ 72.7 \%) & 0.42 ($\uparrow$ 61.5 \%) \\ 
VAP Net         & 0.05 & 0.12 & 0.11 & 0.26 \\ 
\bottomrule
\end{tabular}
\label{abla2}
\end{table*}
Overall, the ADE and FDE metrics across all algorithms were relatively low. In addition to the fact that more than half of the vehicles at a traffic-light-controlled intersection were stationary or low-speed, lane-changing behavior was almost completely absent, resulting in universally lower errors at this intersection. For 12-frame predictions, our proposed method outperforms the others by 43\%, while for 18-frame predictions, our proposed method outperforms the others by over 31\%. As opposed to other methods that rely solely on trajectory information and are less responsive to discontinuous temporal changes \cite{social-gan,socialpooling,salzmann2020trajectron++,rowe2023fjmp}, our model is much more sensitive to traffic light changes. Furthermore, our approach can identify the type and size of vehicles, employing distinct prediction methods for motor vehicles such as cars and non-motorized vehicles like bicycles and motorcycles. This also contributes positively to the improvement of trajectory prediction accuracy.

\textbf{Qualitative Results}

Figure \ref{fig:QualitativeResults} illustrates selected trajectory prediction segments from the validation set. The top three images correspond to 12-frame observations with 12-frame predictions, while the bottom three images pertain to 12-frame observations with 18-frame predictions. Visually, the congruence between predicted trajectories and the ground truth is remarkably high for both sparse and dense traffic scenarios, affirming the low overall ADE and FDE metrics.

\subsection{Ablation Experiments}

In the ablation study, to determine the effectiveness of each component of our model, we systematically removed each encoder except the Trajectory Encoder and observed the model performance. 

The data in Table \ref{abla1} illustrates the significant role each encoder plays in the model's performance. For instance, the exclusion of the Traffic Encoder leads to an increase in ADE by 940.0\% and FDE by 841.7\% for the 12-frame observation and prediction scenario, which underscores the encoder's critical role in capturing traffic light-related dynamics. Similarly, the omission of the Motion Encoder results in a less pronounced but still substantial increase in error rates (ADE by 180.0\% and FDE by 166.7\% for the 12-frame observation and prediction scenario), indicating its importance in understanding vehicular motion. The combined use of all encoders minimizes prediction errors, achieving the lowest ADE and FDE, which confirms the synergistic effect of the integrated information from all encoders on the model's accuracy.

Furthermore, we assessed how various pooling modules affected the overall model performance. According to Figure \ref{abla2}, the results indicate that the pooling approach is extremely important for trajectory prediction accuracy. The study indicated that attention to vehicle dynamics and interactions significantly enhances the accuracy of prediction with our proposed Vehicular Attention Pooling Net (VAP-
Net). Based on Table \ref{abla2}, both the ADE and FDE metrics are lower compared to the Social Pool Net \cite{socialpooling} and Hidden Pool Net \cite{social-gan}.

%% file: sec/5_conclusion.tex
\section{Conclusion}

Our work on the Knowledge-Informed Generative Adversarial Network (KI-GAN) represents a significant stride in trajectory forecasting at signalized intersections. The incorporation of a multi-encoder framework, including the innovative Vehicle Attention Pooling Net (VAP-Net), has proven effective in accurately modeling the complexities of intersection dynamics. A key strength of KI-GAN is the integration of critical traffic signal information and vehicle interactions, which is evident in its performance metrics. Among the key components, the VAP-Net stands out, demonstrating the effectiveness of its attention mechanism in improving trajectory prediction accuracy. As a result of this mechanism's emphasis on vehicle speed and proximity between vehicles, which are crucial in intersection scenarios, the model is able to predict vehicle paths under varying traffic conditions with greater accuracy. 

With the promising results of KI-GAN, further advancements in traffic management systems and autonomous intersection navigation can be achieved, emphasizing that the model can be applied to real-world scenarios. For a more in-depth understanding of interactions with other road users (e.g., vulnerable road users) at intersections, future developments could include integrating more contextual data or refining VAP-Net's attention process. 
In navigating the complexity of urban traffic environments, KI-GAN offers a comprehensive and robust solution, setting a new standard.

%% file: main.bbl
\begin{thebibliography}{49}
\providecommand{\natexlab}[1]{#1}
\providecommand{\url}[1]{\texttt{#1}}
\expandafter\ifx\csname urlstyle\endcsname\relax
  \providecommand{\doi}[1]{doi: #1}\else
  \providecommand{\doi}{doi: \begingroup \urlstyle{rm}\Url}\fi

\bibitem[Abdelraouf et~al.(2022)Abdelraouf, Abdel-Aty, Wang, and Zheng]{abdelraouf2022trajectory}
Amr Abdelraouf, Mohamed Abdel-Aty, Zijin Wang, and Ou Zheng.
\newblock Trajectory prediction for vehicle conflict identification at intersections using sequence-to-sequence recurrent neural networks.
\newblock \emph{arXiv preprint arXiv:2210.08009}, 2022.

\bibitem[Abdelraouf and Han(2023)]{abdelraouf2023interaction}
Rohit Abdelraouf, Amr an d~Gupta and Kyungtae Han.
\newblock Interaction-aware personalized vehicle trajectory prediction using temporal graph neural networks.
\newblock \emph{arXiv preprint arXiv:2308.07439}, 2023.

\bibitem[Alahi et~al.(2016)Alahi, Goel, Ramanathan, Robicquet, Fei-Fei, and Savarese]{socialpooling}
Alexandre Alahi, Kratarth Goel, Vignesh Ramanathan, Alexandre Robicquet, Li Fei-Fei, and Silvio Savarese.
\newblock Social lstm: Human trajectory prediction in crowded spaces.
\newblock In \emph{Proceedings of the IEEE conference on computer vision and pattern recognition}, pages 961--971, 2016.

\bibitem[Altch{\'e} and de~La~Fortelle(2017)]{altche2017lstm}
Florent Altch{\'e} and Arnaud de La~Fortelle.
\newblock An lstm network for highway trajectory prediction.
\newblock In \emph{2017 IEEE 20th international conference on intelligent transportation systems (ITSC)}, pages 353--359. IEEE, 2017.

\bibitem[Bock et~al.(2020)Bock, Krajewski, Moers, Runde, Vater, and Eckstein]{bock2020ind}
Julian Bock, Robert Krajewski, Tobias Moers, Steffen Runde, Lennart Vater, and Lutz Eckstein.
\newblock The ind dataset: A drone dataset of naturalistic road user trajectories at german intersections.
\newblock In \emph{2020 IEEE Intelligent Vehicles Symposium (IV)}, pages 1929--1934. IEEE, 2020.

\bibitem[Breuer et~al.(2020)Breuer, Term{\"o}hlen, Homoceanu, and Fingscheidt]{breuer2020opendd}
Antonia Breuer, Jan-Aike Term{\"o}hlen, Silviu Homoceanu, and Tim Fingscheidt.
\newblock opendd: A large-scale roundabout drone dataset.
\newblock In \emph{2020 IEEE 23rd International Conference on Intelligent Transportation Systems (ITSC)}, pages 1--6. IEEE, 2020.

\bibitem[Cao et~al.(2021)Cao, Zhao, Zeng, Wang, and Long]{Cao2021Real-Time}
Qianxia Cao, Zhongxing Zhao, Qiaoqiong Zeng, Zhengwu Wang, and K. Long.
\newblock Real-time vehicle trajectory prediction for traffic conflict detection at unsignalized intersections.
\newblock \emph{Journal of Advanced Transportation}, 2021.

\bibitem[Chang et~al.(2019)Chang, Lambert, Sangkloy, Singh, Bak, Hartnett, Wang, Carr, Lucey, Ramanan, et~al.]{chang2019argoverse}
Ming-Fang Chang, John Lambert, Patsorn Sangkloy, Jagjeet Singh, Slawomir Bak, Andrew Hartnett, De Wang, Peter Carr, Simon Lucey, Deva Ramanan, et~al.
\newblock Argoverse: 3d tracking and forecasting with rich maps.
\newblock In \emph{Proceedings of the IEEE/CVF conference on computer vision and pattern recognition}, pages 8748--8757, 2019.

\bibitem[Dang et~al.(2023)Dang, Brüdigam, Zhang, Liu, Leibold, and Buss]{Dang2023Distributed}
Ni Dang, Tim Brüdigam, Zengjie Zhang, Fangzhou Liu, Marion Leibold, and Martin Buss.
\newblock Distributed stochastic model predictive control for a microscopic interactive traffic model.
\newblock \emph{Electronics}, 2023.

\bibitem[Dey et~al.(2017)Dey, Martens, Eggen, and Terken]{dey2017impact}
Debargha Dey, Marieke Martens, Berry Eggen, and Jacques Terken.
\newblock The impact of vehicle appearance and vehicle behavior on pedestrian interaction with autonomous vehicles.
\newblock In \emph{Proceedings of the 9th international conference on automotive user interfaces and interactive vehicular applications adjunct}, pages 158--162, 2017.

\bibitem[Eom and Kim(2020)]{eom2020traffic}
Myungeun Eom and Byung-In Kim.
\newblock The traffic signal control problem for intersections: a review.
\newblock \emph{EUROPEAN Transport Research Review}, 12\penalty0 (1), 2020.

\bibitem[Fang et~al.(2022)Fang, Zhang, Zhou, Qian, and Gan]{atten-GAN}
Fang Fang, Pengpeng Zhang, Bo Zhou, Kun Qian, and Yahui Gan.
\newblock Atten-gan: Pedestrian trajectory prediction with gan based on attention mechanism.
\newblock \emph{Cognitive Computation}, 14\penalty0 (6):\penalty0 2296--2305, 2022.

\bibitem[Fayazi and Vahidi(2018)]{Fayazi2018Mixed-Integer}
S.~A. Fayazi and A. Vahidi.
\newblock Mixed-integer linear programming for optimal scheduling of autonomous vehicle intersection crossing.
\newblock \emph{IEEE Transactions on Intelligent Vehicles}, 3:\penalty0 287--299, 2018.

\bibitem[Goodfellow et~al.(2014)Goodfellow, Pouget-Abadie, Mirza, Xu, Warde-Farley, Ozair, Courville, and Bengio]{goodfellow2014generative}
Ian Goodfellow, Jean Pouget-Abadie, Mehdi Mirza, Bing Xu, David Warde-Farley, Sherjil Ozair, Aaron Courville, and Yoshua Bengio.
\newblock Generative adversarial nets.
\newblock \emph{Advances in neural information processing systems}, 27, 2014.

\bibitem[Gupta et~al.(2018)Gupta, Johnson, Fei-Fei, Savarese, and Alahi]{social-gan}
Agrim Gupta, Justin Johnson, Li Fei-Fei, Silvio Savarese, and Alexandre Alahi.
\newblock Social gan: Socially acceptable trajectories with generative adversarial networks.
\newblock In \emph{Proceedings of the IEEE conference on computer vision and pattern recognition}, pages 2255--2264, 2018.

\bibitem[Hochreiter and Schmidhuber(1997)]{lstm}
S. Hochreiter and J. Schmidhuber.
\newblock Long short-term memory.
\newblock \emph{Neural Computation}, 9:\penalty0 1735--1780, 1997.

\bibitem[Islam et~al.(2023)Islam, Abdel-Aty, Goswamy, Abdelraouf, and Zheng]{islam2023effect}
Zubayer Islam, Mohamed Abdel-Aty, Amrita Goswamy, Amr Abdelraouf, and Ou Zheng.
\newblock Effect of signal timing on vehicles’ near misses at intersections.
\newblock \emph{Scientific Reports}, 13\penalty0 (1):\penalty0 9065, 2023.

\bibitem[Kawasaki and Tasaki(2018)]{Kawasaki2018Trajectory}
A. Kawasaki and T. Tasaki.
\newblock Trajectory prediction of turning vehicles based on intersection geometry and observed velocities.
\newblock \emph{2018 IEEE Intelligent Vehicles Symposium (IV)}, pages 511--516, 2018.

\bibitem[Kim et~al.(2017)Kim, Kang, Kim, Lee, Chung, and Choi]{Kim2017Probabilistic}
Byeoungdo Kim, C. Kang, Jaekyum Kim, Seung~Hi Lee, C. Chung, and J. Choi.
\newblock Probabilistic vehicle trajectory prediction over occupancy grid map via recurrent neural network.
\newblock \emph{2017 IEEE 20th International Conference on Intelligent Transportation Systems (ITSC)}, pages 399--404, 2017.

\bibitem[Kosaraju et~al.(2019)Kosaraju, Sadeghian, Mart{\'\i}n-Mart{\'\i}n, Reid, Rezatofighi, and Savarese]{kosaraju2019social}
Vineet Kosaraju, Amir Sadeghian, Roberto Mart{\'\i}n-Mart{\'\i}n, Ian Reid, Hamid Rezatofighi, and Silvio Savarese.
\newblock Social-bigat: Multimodal trajectory forecasting using bicycle-gan and graph attention networks.
\newblock \emph{Advances in neural information processing systems}, 32, 2019.

\bibitem[Krajewski et~al.(2020)Krajewski, Moers, Bock, Vater, and Eckstein]{krajewski2020round}
Robert Krajewski, Tobias Moers, Julian Bock, Lennart Vater, and Lutz Eckstein.
\newblock The round dataset: A drone dataset of road user trajectories at roundabouts in germany.
\newblock In \emph{2020 IEEE 23rd International Conference on Intelligent Transportation Systems (ITSC)}, pages 1--6. IEEE, 2020.

\bibitem[Lee et~al.(2023)Lee, Park, You, Yong, and Moon]{Lee2023Deep}
Seoyoung Lee, Hyogyeong Park, Yeonhwi You, Sungjung Yong, and Il-Young Moon.
\newblock Deep learning-based multimodal trajectory prediction with traffic light.
\newblock \emph{Applied Sciences}, 13\penalty0 (22), 2023.

\bibitem[Li et~al.(2023)Li, Wei, Wu, Barth, Abdelraouf, Gupta, and Han]{li2023personalized}
Siyan Li, Chuheng Wei, Guoyuan Wu, Matthew~J Barth, Amr Abdelraouf, Rohit Gupta, and Kyungtae Han.
\newblock Personalized trajectory prediction for driving behavior modeling in ramp-merging scenarios.
\newblock In \emph{2023 Seventh IEEE International Conference on Robotic Computing (IRC)}, pages 1--4. IEEE, 2023.

\bibitem[Lin et~al.(2022)Lin, Li, Bi, and Qin]{Lin2022Vehicle}
Lei Lin, Weizi Li, Huikun Bi, and Lingqiao Qin.
\newblock Vehicle trajectory prediction using lstms with spatial–temporal attention mechanisms.
\newblock \emph{IEEE Intelligent Transportation Systems Magazine}, 14:\penalty0 197--208, 2022.

\bibitem[Liu et~al.(2021)Liu, Zhang, Fang, Jiang, and Zhou]{liu2021multimodal}
Yicheng Liu, Jinghuai Zhang, Liangji Fang, Qinhong Jiang, and Bolei Zhou.
\newblock Multimodal motion prediction with stacked transformers.
\newblock In \emph{Proceedings of the IEEE/CVF Conference on Computer Vision and Pattern Recognition}, pages 7577--7586, 2021.

\bibitem[Liu et~al.(2022)Liu, Qi, Sisbot, and Oguchi]{liu2022multi}
Yongkang Liu, Xuewei Qi, Emrah~Akin Sisbot, and Kentaro Oguchi.
\newblock Multi-agent trajectory prediction with graph attention isomorphism neural network.
\newblock In \emph{2022 IEEE Intelligent Vehicles Symposium (IV)}, pages 273--279. IEEE, 2022.

\bibitem[Mersch et~al.(2021)Mersch, Höllen, Zhao, Stachniss, and Roscher]{Mersch2021Maneuver-based}
Benedikt Mersch, Thomas Höllen, Kun Zhao, C. Stachniss, and R. Roscher.
\newblock Maneuver-based trajectory prediction for self-driving cars using spatio-temporal convolutional networks.
\newblock \emph{2021 IEEE/RSJ International Conference on Intelligent Robots and Systems (IROS)}, pages 4888--4895, 2021.

\bibitem[Messaoud et~al.(2020)Messaoud, Yahiaoui, Verroust-Blondet, and Nashashibi]{messaoud2020attention}
Kaouther Messaoud, Itheri Yahiaoui, Anne Verroust-Blondet, and Fawzi Nashashibi.
\newblock Attention based vehicle trajectory prediction.
\newblock \emph{IEEE Transactions on Intelligent Vehicles}, 6\penalty0 (1):\penalty0 175--185, 2020.

\bibitem[Mo et~al.(2021)Mo, Xing, and Lv]{Mo2021Graph}
Xiaoyu Mo, Yang Xing, and Chen Lv.
\newblock Graph and recurrent neural network-based vehicle trajectory prediction for highway driving.
\newblock \emph{2021 IEEE International Intelligent Transportation Systems Conference (ITSC)}, pages 1934--1939, 2021.

\bibitem[Mo et~al.(2022)Mo, Huang, Xing, and Lv]{mo2022multi}
Xiaoyu Mo, Zhiyu Huang, Yang Xing, and Chen Lv.
\newblock Multi-agent trajectory prediction with heterogeneous edge-enhanced graph attention network.
\newblock \emph{IEEE Transactions on Intelligent Transportation Systems}, 23\penalty0 (7):\penalty0 9554--9567, 2022.

\bibitem[Navarro and Oh(2022)]{navarro2022social}
Ingrid Navarro and Jean Oh.
\newblock Social-patternn: Socially-aware trajectory prediction guided by motion patterns.
\newblock In \emph{2022 IEEE/RSJ International Conference on Intelligent Robots and Systems (IROS)}, pages 9859--9864. IEEE, 2022.

\bibitem[Oh and Peng(2019)]{Oh2019Impact}
Geunseob Oh and H. Peng.
\newblock Impact of traffic lights on trajectory forecasting of human-driven vehicles near signalized intersections.
\newblock \emph{arXiv: Robotics}, 2019.

\bibitem[Pecher et~al.(2016)Pecher, Hunter, and Fujimoto]{Pecher2016Data-Driven}
P. Pecher, M. Hunter, and R. Fujimoto.
\newblock Data-driven vehicle trajectory prediction.
\newblock \emph{Proceedings of the 2016 ACM SIGSIM Conference on Principles of Advanced Discrete Simulation}, 2016.

\bibitem[Pellegrini et~al.(2009)Pellegrini, Ess, Schindler, and Van~Gool]{pellegrini2009you}
Stefano Pellegrini, Andreas Ess, Konrad Schindler, and Luc Van~Gool.
\newblock You'll never walk alone: Modeling social behavior for multi-target tracking.
\newblock In \emph{2009 IEEE 12th international conference on computer vision}, pages 261--268. IEEE, 2009.

\bibitem[Peng et~al.(2023)Peng, Wu, and Boriboonsomsin]{peng2023energy}
Dongbo Peng, Guoyuan Wu, and Kanok Boriboonsomsin.
\newblock Energy-efficient dispatching of battery electric truck fleets with backhauls and time windows.
\newblock \emph{SAE International Journal of Electrified Vehicles}, 13\penalty0 (14-13-01-0009), 2023.

\bibitem[Quintanar et~al.(2021)Quintanar, Llorca, Parra, Izquierdo, and Sotelo]{Quintanar2021Predicting}
A. Quintanar, D.~F. Llorca, I. Parra, R. Izquierdo, and M. Sotelo.
\newblock Predicting vehicles trajectories in urban scenarios with transformer networks and augmented information.
\newblock \emph{2021 IEEE Intelligent Vehicles Symposium (IV)}, pages 1051--1056, 2021.

\bibitem[Rowe et~al.(2023)Rowe, Ethier, Dykhne, and Czarnecki]{rowe2023fjmp}
Luke Rowe, Martin Ethier, Eli-Henry Dykhne, and Krzysztof Czarnecki.
\newblock Fjmp: Factorized joint multi-agent motion prediction over learned directed acyclic interaction graphs.
\newblock In \emph{Proceedings of the IEEE/CVF Conference on Computer Vision and Pattern Recognition}, pages 13745--13755, 2023.

\bibitem[Roy et~al.(2019)Roy, Ishizaka, Mohan, and Fukuda]{Roy2019Vehicle}
Debaditya Roy, Tetsuhiro Ishizaka, C. Mohan, and A. Fukuda.
\newblock Vehicle trajectory prediction at intersections using interaction based generative adversarial networks.
\newblock \emph{2019 IEEE Intelligent Transportation Systems Conference (ITSC)}, pages 2318--2323, 2019.

\bibitem[Salzmann et~al.(2020)Salzmann, Ivanovic, Chakravarty, and Pavone]{salzmann2020trajectron++}
Tim Salzmann, Boris Ivanovic, Punarjay Chakravarty, and Marco Pavone.
\newblock Trajectron++: Dynamically-feasible trajectory forecasting with heterogeneous data.
\newblock In \emph{Computer Vision--ECCV 2020: 16th European Conference, Glasgow, UK, August 23--28, 2020, Proceedings, Part XVIII 16}, pages 683--700. Springer, 2020.

\bibitem[Wang et~al.(2020)Wang, Khajepour, Cao, and Liu]{wang2020ethical}
Hong Wang, Amir Khajepour, Dongpu Cao, and Teng Liu.
\newblock Ethical decision making in autonomous vehicles: Challenges and research progress.
\newblock \emph{IEEE Intelligent Transportation Systems Magazine}, 14\penalty0 (1):\penalty0 6--17, 2020.

\bibitem[Wei et~al.(2024)Wei, Qin, Wu, Barth, Abdelraouf, Gupta, and Han]{wei2024dilemma}
Chuheng Wei, Ziye Qin, Guoyuan Wu, Matthew~J Barth, Amr Abdelraouf, Rohit Gupta, and Kyungtae Han.
\newblock Dilemma zone: A comprehensive study of influential factors and behavior analysis.
\newblock In \emph{2024 Forum for Innovative Sustainable Transportation Systems (FISTS)}, pages 1--8. IEEE, 2024.

\bibitem[Xu et~al.(2022)Xu, Shao, Li, Yang, Wang, Huang, Lv, and Wang]{sind}
Yanchao Xu, Wenbo Shao, Jun Li, Kai Yang, Weida Wang, Hua Huang, Chen Lv, and Hong Wang.
\newblock Sind: A drone dataset at signalized intersection in china.
\newblock In \emph{2022 IEEE 25th International Conference on Intelligent Transportation Systems (ITSC)}, pages 2471--2478. IEEE, 2022.

\bibitem[Ye et~al.(2021)Ye, Cao, and Chen]{ye2021tpcn}
Maosheng Ye, Tongyi Cao, and Qifeng Chen.
\newblock Tpcn: Temporal point cloud networks for motion forecasting.
\newblock In \emph{Proceedings of the IEEE/CVF Conference on Computer Vision and Pattern Recognition}, pages 11318--11327, 2021.

\bibitem[Zhan et~al.(2018)Zhan, Sun, Hu, Li, and Tomizuka]{Zhan2018Towards}
W. Zhan, Liting Sun, Yeping Hu, Jiachen Li, and M. Tomizuka.
\newblock Towards a fatality-aware benchmark of probabilistic reaction prediction in highly interactive driving scenarios.
\newblock \emph{2018 21st International Conference on Intelligent Transportation Systems (ITSC)}, pages 3274--3280, 2018.

\bibitem[Zhan et~al.(2019)Zhan, Sun, Wang, Shi, Clausse, Naumann, Kummerle, Konigshof, Stiller, de~La~Fortelle, et~al.]{zhan2019interaction}
Wei Zhan, Liting Sun, Di Wang, Haojie Shi, Aubrey Clausse, Maximilian Naumann, Julius Kummerle, Hendrik Konigshof, Christoph Stiller, Arnaud de La~Fortelle, et~al.
\newblock Interaction dataset: An international, adversarial and cooperative motion dataset in interactive driving scenarios with semantic maps.
\newblock \emph{arXiv preprint arXiv:1910.03088}, 2019.

\bibitem[Zhang et~al.(2022)Zhang, Wang, Guo, Lv, Xu, Chen, and Manocha]{zhang2022d2}
Yuzhen Zhang, Wentong Wang, Weizhi Guo, Pei Lv, Mingliang Xu, Wei Chen, and Dinesh Manocha.
\newblock D2-tpred: Discontinuous dependency for trajectory prediction under traffic lights.
\newblock In \emph{European Conference on Computer Vision}, pages 522--539. Springer, 2022.

\bibitem[Zhao et~al.(2021)Zhao, Liu, Al-Dubai, Zomaya, Min, and Hawbani]{Zhao2021A}
Liang Zhao, Yufei Liu, A. Al-Dubai, Albert~Y. Zomaya, G. Min, and Ammar Hawbani.
\newblock A novel generation-adversarial-network-based vehicle trajectory prediction method for intelligent vehicular networks.
\newblock \emph{IEEE Internet of Things Journal}, 8:\penalty0 2066--2077, 2021.

\bibitem[Zhao and Wei(2022)]{zhao2022analysis}
Zhanle Zhao and Chuheng Wei.
\newblock An analysis of the brake performance of radar-based adaptive cruise control during ramp merging on simulation software.
\newblock In \emph{2022 7th International Conference on Intelligent Computing and Signal Processing (ICSP)}, pages 1112--1115. IEEE, 2022.

\bibitem[Zyner et~al.(2018)Zyner, Worrall, and Nebot]{Zyner2018Naturalistic}
Alex Zyner, Stewart Worrall, and E. Nebot.
\newblock Naturalistic driver intention and path prediction using recurrent neural networks.
\newblock \emph{IEEE Transactions on Intelligent Transportation Systems}, 21:\penalty0 1584--1594, 2018.

\end{thebibliography}
